\documentclass[letter, 10 pt, conference]{ieeeconf}  

\IEEEoverridecommandlockouts 
\usepackage[english]{babel}
\usepackage[utf8]{inputenc}
\usepackage{times} 
\usepackage{cite}

\usepackage{booktabs}

\usepackage[hidelinks]{hyperref}
\usepackage[nolist]{acronym}

\usepackage{multirow}

\usepackage{subcaption}
\usepackage{balance}

\usepackage{siunitx}
\usepackage{graphicx} 
\usepackage{epsfig} 
\usepackage{pgfplots}
\pgfplotsset{compat=1.18}
\usepgfplotslibrary{statistics}
\usepgfplotslibrary{groupplots}

\usepackage{float}

\usepackage[export]{adjustbox}

\usepackage{xcolor}
\definecolor{myYellow}{rgb}{0.93,0.69,0.13}
\definecolor{myPurple}{rgb}{0.49,0.18,0.56}
\definecolor{myGreen}{rgb}{0.26 0.72 0.54}
\definecolor{darkgreen}{rgb}{0.272, 0.50, 0.376}
\definecolor{lightgreen}{rgb}{0.585, 0.82, 0.647}

\colorlet{mydarkblue}{blue!30!black}

\usepackage{mathtools}
\usepackage{nicefrac}
\usepackage{amsmath}
\usepackage{amssymb}
\usepackage{bm} 
\usepackage{scalerel} 

\DeclareMathAlphabet{\pazocal}{OMS}{zplm}{m}{n}

\usepackage{etoolbox}
\makeatletter%
\AfterPreamble{%
	\usepackage{hyperref}%
	\let\oldhypertarget\hypertarget%
	\renewcommand{\hypertarget}[2]{%
		\oldhypertarget{#1}{#2}%
		\protected@write\@mainaux{}{%
			\string\expandafter\string\gdef%
			\string\csname\string\detokenize{#1}\string\endcsname{#2}%
		}%
	}%
	\newcommand{\myhyperlink}[1]{%
		\hyperlink{#1}{\csname #1\endcsname}%
	}%
}
\makeatother%

\newcounter{Remark}
\newcounter{Problem}
\usepackage{algorithm}
\usepackage[noend]{algpseudocode}

\makeatletter
\def\BState{\State\hskip-\ALG@thistlm}
\makeatother

\usepackage{tikz}
\pgfdeclarelayer{foreground}
\pgfsetlayers{background, main, foreground}
\usetikzlibrary{quotes, angles, backgrounds, arrows, automata, shapes, positioning, calc, through, spy, decorations.pathreplacing, decorations.markings, arrows.meta, automata, petri, shapes.multipart}

\tikzset{
    imglabel/.style={
      rectangle,
      inner sep=2pt,
      text=black,
      minimum height=1em,
      text centered,
      fill=white,
      fill opacity=1.0,
      text opacity=1,
      anchor=south west,
    },
  }
\tikzset{
	state/.style={
		rectangle,
		draw=black, very thick,
		minimum height=1.0em,
		text centered,
	},
}
\tikzset{
  on each segment/.style={
    decorate,
    decoration={
      show path construction,
      moveto code={},
      lineto code={
        \path [#1]
        (\tikzinputsegmentfirst) -- (\tikzinputsegmentlast);
      },
      curveto code={
        \path [#1] (\tikzinputsegmentfirst)
        .. controls
        (\tikzinputsegmentsupporta) and (\tikzinputsegmentsupportb)
        ..
        (\tikzinputsegmentlast);
      },
      closepath code={
        \path [#1]
        (\tikzinputsegmentfirst) -- (\tikzinputsegmentlast);
      },
    },
  },
  mid arrow/.style={postaction={decorate,decoration={
        markings,
        mark=at position .5 with {\arrow[#1]{stealth}}
      }}},
}

\tikzset{
  half circle/.style={
      semicircle,
      shape border rotate=180,
      anchor=chord center,
      minimum size=5mm
      }
}

\graphicspath{{./figures/}}

\newcommand\copyrighttext{%
    \small \begin{center} \color{red} \textcopyright\,2026 The Authors. This work has been accepted for presentation at the ICRA 2026 Workshop on ``Aerial inspection for marine infrastructures," June 1, 2026, Vienna, Austria. Personal use of this material is permitted. For all other uses, including reprinting, republishing, redistribution to servers or mailing lists, resale, inclusion in collective works, or reuse of any copyrighted component of this work in other works, permission must be obtained from the copyright holder(s) \end{center}}
\newcommand\copyrightnotice{%
	\begin{tikzpicture}[remember picture,overlay]
	\node[anchor=south,yshift=25.6cm] at (current page.south) 
	{\color{red}\fbox{\parbox{\dimexpr\textwidth-\fboxsep-\fboxrule\relax}{\copyrighttext}}};
	\end{tikzpicture}%
}

\title{\copyrightnotice \LARGE \bf Sensitivity-Based Robust NMPC for Close-Proximity Offshore Wind Turbine Inspection with a Tilted Multirotor}

\author{Giuseppe Silano$^{1}$ and Martin Saska$^{2}$  
    \thanks{Partially funded by the Italian Electrical System research fund (decree n.~388, Nov.~6, 2024), GA\v{C}R project no.~26-22419S, CTU grant no.~SGS26/077/OHK3/1T/13, and the European project ``Robotics and Advanced Industrial Production'' (no.~CZ.02.01.01/00/22008/0004590).
    $^1$G.~Silano is with the Ricerca sul Sistema Energetico S.p.A., Milan, Italy, and the Czech Technical University in Prague, Czechia ({\tt\small giuseppe.silano@fel.cvut.cz}).
    $^2$M.~Saska is with the Czech Technical University in Prague, Czechia ({\tt\small martin.saska@cvut.cz}).}
}

\begin{document}

\maketitle
\thispagestyle{empty} 
\pagestyle{empty} 


\begin{acronym}
    \acro{GTMR}[GTMR]{Generically-Tilted Multirotor}
    \acro{MPC}[MPC]{Model Predictive Control}
    \acro{MRAV}[MRAV]{Multi-Rotor Aerial Vehicle}
    \acro{NLP}[NLP]{Nonlinear Programming}
    \acro{NMPC}[NMPC]{Nonlinear Model Predictive Control}
    \acro{RMSE}[RMSE]{Root Mean Square Error}
    \acro{UAV}[UAV]{Unmanned Aerial Vehicle}
    \acro{wrt}[w.r.t.]{with respect to}
\end{acronym}



\begin{abstract}
    Close-proximity offshore wind turbine inspection requires strict clearance control around large cylindrical structures under wind and model mismatch. Nominal \ac{NMPC} may violate safety constraints when mass, inertia, thrust effectiveness, drag, or wind conditions differ from nominal assumptions. We propose a sensitivity-based robust NMPC for a tilted multirotor that robustifies the tower-clearance constraint via online constraint tightening. First-order parametric state sensitivities provide a structured-uncertainty margin, while bounded gusts are handled by a stage-dependent additive margin. The formulation augments the nominal \ac{NMPC} with sensitivity propagation and margin evaluation only, leaving the receding-horizon optimization structure unchanged. Monte-Carlo evaluation over $500$ uncertainty realizations on a boundary-critical helical inspection trajectory shows that the proposed controller eliminates the clearance violations observed under nominal \ac{NMPC} at the cost of a moderate increase in solve time.
\end{abstract}






\section{Introduction}
\label{sec:introduction}

Aerial inspection can reduce offshore maintenance cost and human exposure, but reliable autonomy near wind-turbine towers remains difficult because of persistent wind, aerodynamic effects, and plant--model mismatch \cite{Liu2022RSER,MoolanFeroze2019ICRA,CarIEEEAccess2020}. Since inspection-quality sensing requires tracking close-proximity trajectories around the tower while enforcing a strict minimum clearance, safety preservation rather than trajectory tracking alone becomes the primary control objective.

\acf{NMPC} is well suited to this setting because it handles nonlinear dynamics and constraints within a unified receding-horizon problem \cite{Bicego2020JIRS, Findeisen2003NMPCOverview}. However, nominal \ac{NMPC} may be overly optimistic under uncertainty, yielding trajectories that are feasible in prediction but unsafe on the real system. This paper therefore proposes a practical robustification for a tilted multirotor based on sensitivity-driven tightening of the cylindrical clearance constraint under bounded structured uncertainty and bounded wind gusts.

Robustifications of \ac{NMPC} have been extensively studied in the process and robotics control communities. Min-max formulations explicitly optimize against worst-case disturbance realizations but are generically computationally prohibitive for fast aerial platforms~\cite{Mayne2014survey}. Multi-stage and scenario-based \ac{NMPC} mitigate this cost by branching the prediction tree over a finite set of uncertainty realizations~\cite{Lucia2014}, but the combinatorial growth of scenarios remains demanding for the state dimensions and sample rates typical of multirotors. Tube-based \ac{NMPC}, in either rigid or homothetic form, decouples a nominal trajectory from an ancillary feedback law that keeps the perturbed state inside a robust positive invariant set~\cite{Giordano2018ICRA, BelvedereTRO2025}; for tilted multirotors, the joint presence of structured parametric mismatch (mass, inertia, thrust and drag effectiveness) and exogenous wind makes the construction of such invariant tubes both restrictive and conservative. Sensitivity-based constraint tightening, instead, propagates first-order parametric sensitivities along the prediction horizon and uses them to inflate the constraints on the fly~\cite{Zanelli2021,Houska2015}. The proposed controller belongs to this last family, specialized to the cylindrical clearance constraint relevant for close-proximity tower inspection and complemented by a closed-form gust margin that does not enter the sensitivity propagation.

More specifically, the contribution is threefold: i) a structured uncertainty model capturing mismatch in mass, inertia, propulsion effectiveness, drag, and persistent wind bias; ii) an online first-order parametric tightening term obtained from propagated state sensitivities; and iii) an additive stage-dependent gust margin in explicit closed form.



\section{Problem Setup and Prediction Model}
\label{sec:problem_setup}

This paper considers close-proximity helical tower inspection with guaranteed minimum clearance. The tower is modeled as a vertical cylinder defining both the inspection geometry and safety-critical clearance constraint, while the aerial platform uses a control-oriented tilted-multirotor prediction model with input-rate actuation.



\subsection{Wind turbine inspection geometry}
\label{sec:windTurbineInspectionGeomtry}

The tower is modeled as a vertical cylinder of radius $R$, aligned with the inertial $z$-axis, which captures the dominant obstacle during close-proximity tower inspection. Let $\mathbf{p}=[x\ y\ z]^\top$ denote the vehicle position, and define $\rho(\mathbf{p})\triangleq \sqrt{x^2+y^2}$ and $d_T(\mathbf{p})\triangleq \rho(\mathbf{p})-R$. Collision avoidance requires \vspace{-1em}

%
\begin{equation}\label{eq:collisionAvoidanceConstraint}
    d_T(\mathbf{p}(t))\ge d_{\min}, \quad \forall t\ge 0,
\end{equation}
where $d_{\min}>0$ is the minimum admissible clearance. The inspection reference is a helical trajectory $\mathbf{p}_\mathrm{ref}$ on the preferred offset surface
\begin{equation}\label{eq:preferredDistance}
    d_T(\mathbf{p}_{\mathrm{ref}}(t)) = d_{\mathrm{ref}} > d_{\min},
\end{equation}
so that the vehicle orbits the tower while maintaining the desired sensing distance.



\subsection{Control-oriented GTMR model}
\label{sec:gtmrModel}

The aerial platform is described by a control-oriented \ac{GTMR} model \cite{HamandiIJRR2021,Ryll2019IJRR} with input-rate actuation. Compared with collinear designs, the \ac{GTMR} allocation provides non-zero lateral thrust components, which (i) partially decouple position and attitude control, allowing the vehicle to maintain a tower-facing orientation along the helical orbit without pitching toward the structure, and (ii) increase the admissible thrust direction set, improving rejection of lateral wind gusts during close-proximity flight. These features make the \ac{GTMR} allocation particularly suitable for the safety-critical inspection scenario considered in this paper. Let $N_p$ denote the number of rotor--propeller units, and define the system state as $\mathbf{x}= [\mathbf{p}^\top\ \boldsymbol{\eta}^\top\ \mathbf{v}^\top\ \boldsymbol{\omega}^\top\ \boldsymbol{\xi}^\top]^\top,$
%
%
where $\boldsymbol{\eta} = [\varphi\ \vartheta\ \psi]^\top \in \mathbb{R}^3$ is the roll--pitch--yaw Euler-angle attitude parametrization, used here as a local representation valid away from the gimbal-lock singularities encountered in close-proximity orbital flight, $\mathbf{v}\in \mathbb{R}^3$ is the linear velocity expressed in the inertial frame, $\boldsymbol{\omega} \in \mathbb{R}^3$ is the body angular velocity, $\boldsymbol{\xi} \in \mathbb{R}^{N_p}$ collects the rotor-speed actuator states,
and $\mathbf{u} = \dot{\boldsymbol{\xi}} \in \mathbb{R}^{N_p}$ is the actuator-rate input. The prediction model is written compactly as
\begin{equation}
    \dot{\mathbf{x}} = f(\mathbf{x},\mathbf{u},\boldsymbol{\zeta})+\mathbf{G}_w\mathbf{w}_g,
    \label{eq:gtmr_dynamics}
\end{equation}
where $\boldsymbol{\zeta}$ collects the structured uncertainty, including the slowly varying wind-induced force bias, $\mathbf{w}_g$ is the bounded fast-varying gust force acting on the airframe, and $\mathbf{G}_w$ is a constant selection matrix mapping gusts to the translational acceleration channels of the state dynamics. The state and input are constrained to the admissible sets $\mathbf{x}\in\mathcal{X}$ and $\mathbf{u}\in\mathcal{U}$, which encode actuator magnitude and rate limits.



\section{Uncertainty Description}
\label{sec:uncertainty_description}

The uncertainty considered in this paper is split into two parts: a slowly varying structured mismatch and a fast bounded gust component. This separation matches the robustification strategy developed later, where structured uncertainty is handled through sensitivity propagation and gusts through an additive margin.




The structured uncertainty vector is
\begin{equation}\label{eq:uncertainVector}
    \boldsymbol{\zeta}=
    \big[
    \delta_m\ \delta_{J_x}\ \delta_{J_y}\ \delta_{J_z}\ \delta_T\ \delta_D\ w_{b,x}\ w_{b,y}\ w_{b,z}
    \big]^\top,
\end{equation}
where $\delta_m$, $\delta_{J_x}$, $\delta_{J_y}$, and $\delta_{J_z}$ denote relative mismatch in mass and principal inertia components, $\delta_T$ and $\delta_D$ represent relative mismatch in thrust effectiveness and drag, and $w_{b,x},w_{b,y},w_{b,z}$ are the components of the persistent wind bias. These quantities enter the prediction model as bounded perturbations of the corresponding nominal parameters, e.g., through multiplicative relations of the form $m=(1+\delta_m)m_0$. The admissible set is defined as $\mathcal{Z}=\{\boldsymbol{\zeta} \in \mathbb{R}^{n_\zeta}:\underline{\boldsymbol{\zeta}}\le\boldsymbol{\zeta}\le\bar{\boldsymbol{\zeta}}\},$
%
%
where the inequalities are understood componentwise, and $\underline{\boldsymbol{\zeta}}$, $\bar{\boldsymbol{\zeta}} \in \mathbb{R}^{n_\zeta}$, with $n_\zeta = 9$, denote the lower and upper bounds on the uncertainty vector.




The aerodynamic force induced by wind is decomposed as $\mathbf{w}(t)=\mathbf{w}_b+\mathbf{w}_g(t)$, where $\mathbf{w}_b$ is a slowly varying bias force included in $\boldsymbol{\zeta}$, and $\mathbf{w}_g(t)$ is a fast-varying bounded gust force satisfying $\|\mathbf{w}_g(t)\|\le \bar{w}_g$ for all $t\ge 0$. 



\section{Nominal NMPC Formulation}
\label{sec:nominal_nmpc}

With sampling time $T_s$, horizon $N$, and nominal uncertainty realization $\boldsymbol{\zeta}_0=\mathbf{0}$, the discrete-time model reads $\mathbf{x}_{i+1}=F(\mathbf{x}_i,\mathbf{u}_i,\boldsymbol{\zeta}_0)$. The nominal \ac{NMPC} problem, used as the baseline for the subsequent robust tightening, is \vspace{-1.40em}
%

%
\begin{subequations}\label{eq:nominalNMPC}
    \begin{align}
    \min_{\{\mathbf{x}_i,\mathbf{u}_i\}}\;&\sum_{i=0}^{N-1}\ell(\mathbf{x}_i,\mathbf{u}_i,\mathbf{r}_i)+\ell_f(\mathbf{x}_N,\mathbf{r}_N) \label{eq:nominalNMPC_cost}\\
    \text{s.t.}\;&\mathbf{x}_{i+1}=F(\mathbf{x}_i,\mathbf{u}_i,\boldsymbol{\zeta}_0),\ i=0,\dots,N-1, \label{eq:nominalNMPC_dyn}\\
    &\mathbf{x}_i\in\mathcal{X},\ \mathbf{u}_i\in\mathcal{U},\ i=0,\dots,N-1, \label{eq:nominalNMPC_bounds}\\
    &d_T(\mathbf{p}_i)\ge d_{\min},\ i=0,\dots,N, \label{eq:nominalNMPC_clearance}\\
    &\mathbf{x}_0=\mathbf{x}(t_k). \label{eq:nominalNMPC_init}
    \end{align}
\end{subequations}

The stage and terminal costs are chosen as
%
\begin{equation}
    \resizebox{0.90\linewidth}{!}{$
        \ell=\|\mathbf{h}(\mathbf{x}_i)-\mathbf{r}_i\|_{\mathbf{Q}}^2+\|\mathbf{u}_i\|_{\mathbf{Q}_u}^2,\;
        \ell_f=\|\mathbf{h}_f(\mathbf{x}_N)-\mathbf{r}_N\|_{\mathbf{Q}_f}^2.
    $}
    \label{eq:costs}
\end{equation}

The reference $\mathbf{r}_i$ samples the helical trajectory and includes desired position, velocity, and acceleration. The output map $\mathbf{h}(\mathbf{x}) = [\mathbf{p}^\top\ \mathbf{v}^\top\ \dot{\mathbf{v}}^\top(\mathbf{x},\mathbf{u})]^\top$ selects the tracked channels, with $\dot{\mathbf{v}}$ evaluated along~\eqref{eq:gtmr_dynamics}; $\mathbf{h}_f(\mathbf{x}) = \mathbf{p}$. The block-diagonal weight $\mathbf{Q} = \mathrm{blkdiag}(\mathbf{Q}_p, \mathbf{Q}_v, \mathbf{Q}_a)$ uses the values in Table~\ref{tab:simulation_params}; $\mathbf{Q}_u$ regularizes the actuator-rate input and $\mathbf{Q}_f$ the terminal position error.




\section{Sensitivity-Based Tube Tightening}
\label{sec:tightening}

The proposed robustification augments the nominal clearance constraint with a parametric margin for structured mismatch and a gust margin for fast wind disturbances, enforcing a conservative clearance condition along the horizon while preserving a tractable optimization structure.



\subsection{Parametric sensitivity dynamics}

To quantify how structured uncertainty affects the predicted motion, the proposed method propagates first-order state sensitivities along the horizon. To this end, define the state sensitivity matrix as
\begin{equation}
    \mathbf{\Pi}\triangleq \frac{\partial \mathbf{x}}{\partial \boldsymbol{\zeta}}\in\mathbb{R}^{(12+N_p)\times 9},
    \label{eq:PiDef}
\end{equation}
which measures the sensitivity of the extended state to perturbations of the structured uncertainty vector. Its continuous-time evolution, evaluated along the nominal predicted trajectory $(\mathbf{x}_i, \mathbf{u}_i, \boldsymbol{\zeta}_0)$, is
\begin{equation}
    \dot{\mathbf{\Pi}}=\frac{\partial f}{\partial \mathbf{x}}\mathbf{\Pi}+\frac{\partial f}{\partial \boldsymbol{\zeta}},\qquad \mathbf{\Pi}(t_k)=\mathbf{0},
    \label{eq:sensitivityODE}
\end{equation}
where the initialization $\mathbf{\Pi}(t_k)=\mathbf{0}$ reflects the reset of the sensitivity at the beginning of each prediction horizon. The corresponding discrete-time propagation $\mathbf{\Pi}_{i+1}=F_{\Pi}(\mathbf{x}_i,\mathbf{u}_i,\mathbf{\Pi}_i,\boldsymbol{\zeta}_0)$ is obtained consistently with the nominal discrete-time model.

The sensitivity propagation introduces $n_\zeta n_x = 9(12+N_p)$ additional scalar states per stage with respect to the nominal \ac{NMPC}, evaluated at negligible marginal cost via the same automatic differentiation toolchain.\footnote{For the \ac{GTMR} considered in this paper ($N_p=6$), this amounts to $162$ states per stage and $3240$ states over the horizon, translating to the approximately twofold solve-time increase reported in Section~\ref{sec:simulationResults}.}



\subsection{Clearance constraint in standard form}

To derive the tightening term, the clearance requirement is rewritten as a scalar inequality. Specifically, define $y(\mathbf{x})\triangleq d_{\min}-d_T(\mathbf{p})\le 0,$
%
so that satisfying $y(\mathbf{x})\le 0$ is equivalent to enforcing the minimum-clearance condition. Since the constraint depends only on the position components of the state, its Jacobian \ac{wrt} the system state reads
\begin{equation}
    J_{yx}(\mathbf{x})=\frac{\partial y}{\partial \mathbf{x}}=\begin{bmatrix}-x/\rho & -y/\rho & 0 & \mathbf{0}_{1\times(9+N_p)}\end{bmatrix},
    \label{eq:Jyx}
\end{equation}
where the trailing zero block reflects the absence of direct dependence on the remaining state components.
The gradient $\partial y / \partial p$ is well defined throughout the inspection mission since the helical reference enforces $\rho(\mathbf{p}) \ge R + d_{\min} > 0$ at all times. Therefore, the first-order sensitivity of the scalar clearance constraint \ac{wrt} the structured uncertainty reads $\mathbf{\Pi}_y\triangleq\partial y / \partial \boldsymbol{\zeta} = J_{yx}(\mathbf{x})\mathbf{\Pi}$, the key ingredient for the parametric tightening term.



\subsection{Parametric tightening margin}

To account for bounded structured uncertainty, consider the deviation of the uncertainty vector from its nominal value,
\begin{equation}
    \mathbf{w}_{\zeta}\triangleq \boldsymbol{\zeta}-\boldsymbol{\zeta}_0,
\end{equation}
which lies in the componentwise-bounded set
\begin{equation}
    \mathcal{W}_{\zeta}\triangleq
    \left\{
    \mathbf{w}_{\zeta} \in \mathbb{R}^{n_\zeta}:
    |w_{\zeta,j}|\le \bar{w}_{\zeta,j},\
    j=1,\dots,n_\zeta
    \right\},
    \label{eq:Wzeta}
\end{equation}  
where $n_\zeta = 9$ and $\bar{w}_{\zeta,j} > 0$ denotes the componentwise upper bound on the $j$-th uncertainty channel. Using the first-order approximation $\Delta y_i \approx \mathbf{\Pi}_{y,i}\,\mathbf{w}_\zeta$ and exploiting the box structure of $\mathcal{W}_\zeta$, the worst-case clearance perturbation at stage $i$ admits a closed form as the support function of $\mathcal{W}_\zeta$ evaluated at $\mathbf{\Pi}_{y,i}$, attained by sign-aligning each component of $\mathbf{w}_\zeta$ with the corresponding sensitivity. The parametric tightening margin is therefore defined as
\begin{equation}
    \alpha_{p,i} \triangleq \sup_{\mathbf{w}_\zeta \in \mathcal{W}_\zeta}\!\left| \mathbf{\Pi}_{y,i}\,\mathbf{w}_\zeta \right| + \varepsilon_s
    = \sum_{j=1}^{n_\zeta} |\Pi_{y,i,j}|\, \bar{w}_{\zeta,j} + \varepsilon_s,
    \label{eq:alpha_p}
\end{equation}

%
where $\Pi_{y,i,j}$ denotes the $j$-th component of $\Pi_{y,i}$ and $\varepsilon_s > 0$ is a small slack term introduced for numerical robustness. Equation~\eqref{eq:alpha_p} is tighter than the H\"older upper bound $\|\Pi_{y,i}\|_\infty \|\bar{\mathbf{w}}_\zeta\|_1$ whenever the per-channel sensitivities differ in magnitude, which is typical here because $\boldsymbol{\zeta}$ aggregates parameters of heterogeneous physical scales (relative mass, inertia, propulsion, and drag mismatches, and wind-bias forces). Computing~\eqref{eq:alpha_p} requires only a weighted $\ell_1$ norm of $\Pi_{y,i}$ and therefore preserves the per-stage cost of the nominal \ac{NMPC}.



\subsection{Wind-gust margin}

The gust component is handled through an additive stage-dependent margin instead of being included in the sensitivity propagation. Since $m\ge m_{\min}\triangleq(1-\bar{\delta}_m)m_0$, the bound $\|\mathbf{w}_g(t)\|\le \bar{w}_g$ implies $\|\Delta\dot{\mathbf{v}}\|\le \bar{a}_g\triangleq \bar{w}_g / m_{\min}$. Assuming identical initial states for the nominal and perturbed trajectories at the start of the horizon and a worst-case open-loop double integration of the gust acceleration over the prediction window---neglecting any in-horizon corrective action---the position deviation over $\tau=iT_s$ satisfies $\|\Delta\mathbf{p}_g(\tau)\|\le \tfrac{1}{2}\bar{a}_g\tau^2$, which yields the stage-dependent gust margin
\begin{equation}
    \alpha_{g,i}\triangleq \tfrac{1}{2}\bar{a}_g(iT_s)^2.
    \label{eq:gustMargin}
\end{equation}

Combining the parametric and gust margins yields the tightened clearance condition
\begin{equation}
    d_T(\mathbf{p}_i)\ge d_{\min}+\alpha_{p,i}+\alpha_{g,i},
    \label{eq:tightenedConstraintClearance}
\end{equation}
or equivalently $y(\mathbf{x}_i)+\alpha_{p,i}+\alpha_{g,i}\le 0$.

\section{Robustified NMPC Problem}

The proposed controller augments the nominal \ac{NMPC} formulation in Section \ref{sec:nominal_nmpc} with sensitivity propagation and replaces the nominal clearance constraint with its tightened counterpart. The resulting finite-horizon problem is

\vspace{-1em}
\small
\begin{subequations}
    \label{eq:tubeNMPC}
    \begin{align}
    \min_{\{\mathbf{x}_i,\mathbf{u}_i,\mathbf{\Pi}_i\}}\;&\sum_{i=0}^{N-1}\ell(\mathbf{x}_i,\mathbf{u}_i,\mathbf{r}_i)+\ell_f(\mathbf{x}_N,\mathbf{r}_N) \label{eq:tubeNMPC_cost}\\
    \text{s.t.}\;&\mathbf{x}_{i+1}=F(\mathbf{x}_i,\mathbf{u}_i,\boldsymbol{\zeta}_0),\ i=0,\dots,N-1, \label{eq:tubeNMPC_dyn}\\
    &\mathbf{\Pi}_{i+1}=F_{\Pi}(\mathbf{x}_i,\mathbf{u}_i,\mathbf{\Pi}_i,\boldsymbol{\zeta}_0),\ i=0,\dots,N-1, \label{eq:tubeNMPC_sens}\\
    &\mathbf{x}_i\in\mathcal{X},\ \mathbf{u}_i\in\mathcal{U},\ i=0,\dots,N-1, \label{eq:tubeNMPC_bounds}\\
    &y(\mathbf{x}_i)+\alpha_{p,i}+\alpha_{g,i}\le 0,\ i=0,\dots,N, \label{eq:tubeNMPC_tight}\\
    &\mathbf{x}_0=\mathbf{x}(t_k),\quad \mathbf{\Pi}_0=\mathbf{0}. \label{eq:tubeNMPC_init}
    \end{align}
\end{subequations}
\normalsize

Here, $\mathbf{\Pi}_i$ propagates the first-order sensitivity to structured uncertainty, while $\alpha_{p,i}$ and $\alpha_{g,i}$ denote the corresponding parametric and gust tightening margins at prediction stage $i$. 



\section{Simulation Results}
\label{sec:simulationResults}

This section presents the numerical assessment of the proposed robustified \ac{NMPC} in a safety-critical tower-inspection scenario. The mission consists of tracking a helical reference trajectory around a cylindrical tower while enforcing the minimum-clearance requirement \(d_T(\mathbf{p}) \ge d_{\min}\). The numerical values used in the simulations, including platform parameters, inspection geometry, symmetric uncertainty bounds, and controller weights, are summarized in Table~\ref{tab:simulation_params}. The desired inspection distance \(d_{\mathrm{ref}}\) is selected close to \(d_{\min}\), so that moderate uncertainty can erode the clearance margin if it is not explicitly accounted for. Both the nominal and the robustified \ac{NMPC} are implemented with the same prediction horizon, discretization, weighting matrices, actuator constraints, and solver settings to ensure a fair comparison. 
The continuous-time dynamics are discretized by a fixed-step fourth-order Runge--Kutta scheme within a multiple-shooting formulation, and the resulting nonlinear programs are solved in MATLAB using MATMPC \cite{Chen2019ECC} in real-time iteration mode with qpOASES \cite{Ferreau2014}. 

\begin{table}[t]
    \centering
    \caption{Simulation parameters.}
    \vspace{-1em}
    \label{tab:simulation_params}
    \renewcommand{\arraystretch}{1.0}
    \setlength{\tabcolsep}{4pt}
    \begin{tabular}{@{}llcl@{}}
        \toprule
        \textbf{Parameter} & \textbf{Symbol} & \textbf{Value} & \textbf{Unit} \\
        \midrule
        \multicolumn{4}{@{}l}{\textbf{\ac{GTMR} model}} \\
        Number of rotors & $N_p$ & $6$ & --\\
        Rotor tilt angle & $\alpha$ & $20$ & deg \\
        Mass & $m$ & 2.57 & kg \\
        Principal inertia & $\mathbf{J}$ & $\mathrm{diag}(0.11,\,0.11,\,0.19)$ & kg\,m$^2$ \\
        Arm length & $d$ & 0.39 & m \\
        Rotor force coefficient & $c_f$ & $11.8 \times 10^{-4}$ & N/Hz$^2$ \\
        Rotor torque coefficient & $c_t$ & $2.5 \times 10^{-5}$ & N\,m/Hz$^2$ \\
        \midrule
        \multicolumn{4}{@{}l}{\textbf{Inspection geometry}} \\
        Tower radius & $R$ & 1.00 & m \\
        Minimum clearance & $d_{\min}$ & 0.20 & m \\
        Desired offset & $d_{\mathrm{ref}}$ & 0.35 & m \\
        \midrule
        \multicolumn{4}{@{}l}{\textbf{Uncertainty bounds}} \\
        Mass mismatch & $\bar{\delta}_m$ & 0.10 & -- \\
        Inertia mismatch & $\bar{\boldsymbol{\delta}}_J$ & $[0.15,\;0.15,\;0.20]$ & -- \\
        Propulsion mismatch & $\bar{\delta}_T$ & 0.10 & -- \\
        Drag mismatch & $\bar{\delta}_D$ & 0.20 & -- \\
        Wind-bias bounds & $\bar{\mathbf{w}}_b$ & $[0.8,\;0.8,\;0.3]^\top$ & N \\
        Gust-force bound & $\bar{w}_g$ & 0.6 & N \\
        \midrule
        \multicolumn{4}{@{}l}{\textbf{NMPC weights}} \\
        Position tracking & $\mathbf{Q}_p$ & $\mathrm{diag}(50,\,50,\,80)$ & -- \\
        Velocity tracking & $\mathbf{Q}_v$ & $\mathrm{diag}(5,\,5,\,8)$ & -- \\
        Acceleration tracking & $\mathbf{Q}_a$ & $\mathrm{diag}(1,\,1,\,2)$ & -- \\
        Terminal position tracking & $\mathbf{Q}_f$ & $\mathrm{diag}(80,\,80,\,120)$ & -- \\
        \midrule
        \multicolumn{4}{@{}l}{\textbf{Simulation setup}} \\
        Prediction horizon & $N$ & 20 & -- \\
        Sampling time & $T_s$ & 0.025 & s \\
        Simulation time & $T_{\mathrm{sim}}$ & 50 & s \\
        \bottomrule
    \end{tabular}
\end{table}

\begin{figure}[t]
    \centering
    \adjincludegraphics[
        trim={{.21\width} {.07\height} {.21\width} {.10\height}},
        clip,
        width=0.8\linewidth,
        keepaspectratio
    ]{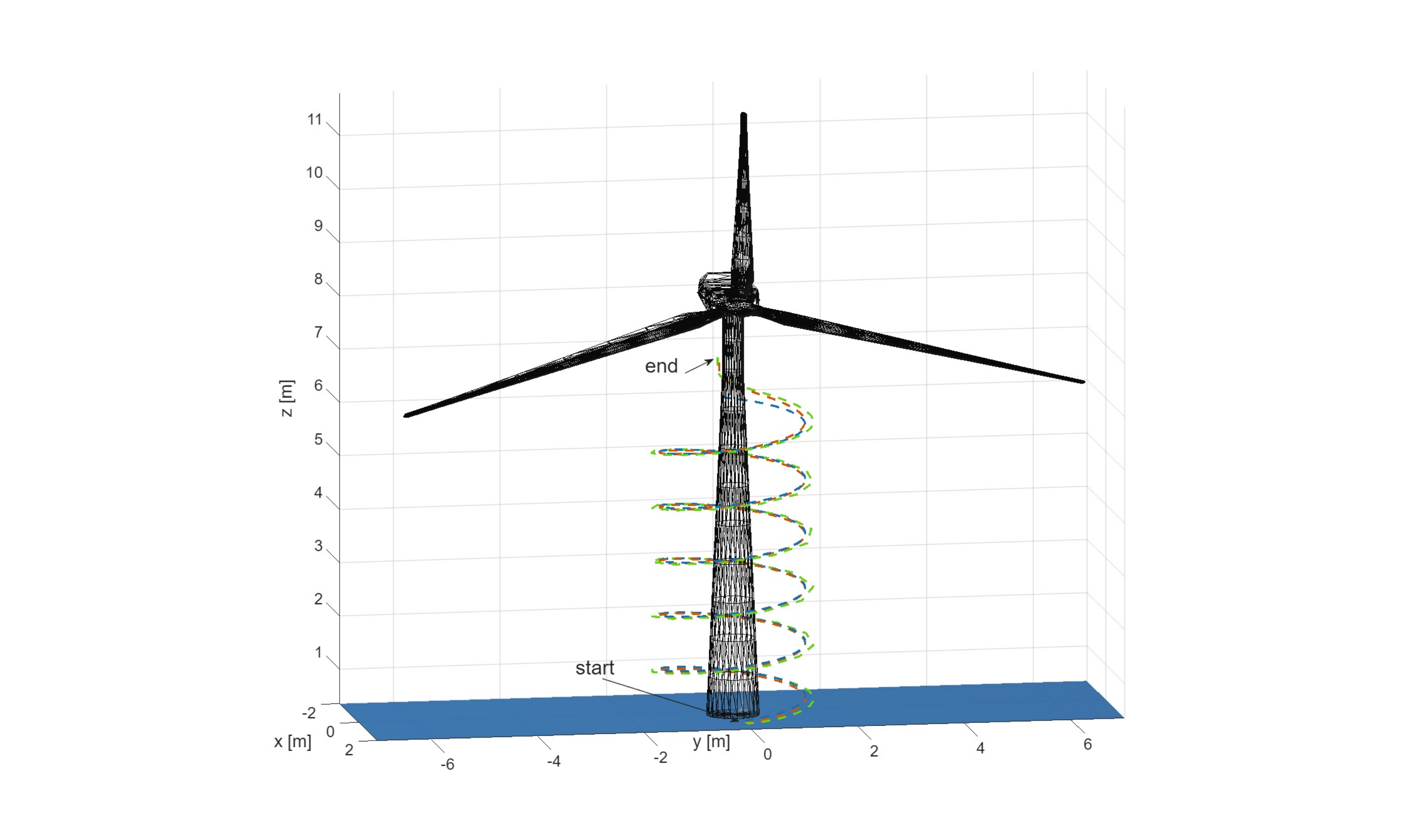}
    \caption{Representative closed-loop trajectories under a wind bias $\mathbf{w}_b = [0.6, 0.4, 0.0]^\top$N: helical reference (blue), nominal \ac{NMPC} (red), and robustified \ac{NMPC} (green). The robustified controller anticipates the bias-induced drift and preserves a larger clearance from the tower.}
    \label{fig:traj_compare}
\end{figure}

\begin{table}[t]
    \centering
    \caption{Comparison between nominal and robustified \ac{NMPC} over 500 uncertainty realizations.}
    \vspace{-1em}
    \label{tab:results_summary}
    \resizebox{\columnwidth}{!}{%
    \begin{tabular}{lccc}
        \toprule
        \textbf{Controller} & \textbf{Min. clearance} & \textbf{Trials with violation} & \textbf{Avg. solve time} \\
        \midrule
        Nominal \ac{NMPC} & 0.182 [m] & 230/500 & 9.4 [ms] \\
        Robustified \ac{NMPC} & 0.364 [m] & 0/500 & 18.7 [ms] \\
        \bottomrule
    \end{tabular}
    }
\end{table}

\begin{figure}[t]
    \centering
    \begin{tikzpicture}
        \scalebox{0.95}{
        \begin{axis}[
        width=2.2119in,
        height=1.8183in,
        at={(0.758in,0.481in)},
        scale only axis,
        unbounded coords=jump,
        xmin=0.5,
        xmax=2.5,
        xtick={1,2},
        xticklabels={{Nominal},{Robust}},
        ymin=-0.3,
        ymax=0.05,
        ylabel={Constraint residual $s(t)$ [m]},
        ylabel style={yshift=-0.255cm, xshift=0cm,font=\color{white!15!black}},
        axis background/.style={fill=white},
        xmajorgrids,
        ymajorgrids,
        ]

        \addplot [color=red, line width=1.0pt, dashed, forget plot]
          coordinates {(0.5, 0) (2.5, 0)};

        \addplot [color=black, dashed, forget plot]
          coordinates {(1, 0.018) (1, -0.08)};
        \addplot [color=black, dashed, forget plot]
          coordinates {(1, -0.22) (1, -0.25)};
        \addplot [color=black, forget plot] coordinates {(0.925, 0.018) (1.075, 0.018)};
        \addplot [color=black, forget plot] coordinates {(0.925, -0.25) (1.075, -0.25)};
        \addplot [color=blue, forget plot]
          coordinates {(0.85, -0.22) (1.15, -0.22) (1.15, -0.08) (0.85, -0.08) (0.85, -0.22)};
        \addplot [color=red, forget plot] coordinates {(0.85, -0.15) (1.15, -0.15)};

        \addplot [color=black, dashed, forget plot]
          coordinates {(2, -0.114) (2, -0.16)};
        \addplot [color=black, dashed, forget plot]
          coordinates {(2, -0.24) (2, -0.28)};
        \addplot [color=black, forget plot] coordinates {(1.925, -0.114) (2.075, -0.114)};
        \addplot [color=black, forget plot] coordinates {(1.925, -0.28) (2.075, -0.28)};
        \addplot [color=blue, forget plot]
          coordinates {(1.85, -0.24) (2.15, -0.24) (2.15, -0.16) (1.85, -0.16) (1.85, -0.24)};
        \addplot [color=red, forget plot] coordinates {(1.85, -0.19) (2.15, -0.19)};

        \end{axis}
        }
    \end{tikzpicture}
    \caption{Signed clearance residuals \(s(t)=d_{\min}-d_T(\mathbf{p}(t))\) over $500$ Monte-Carlo trials; $s(t)\le 0$ indicates constraint satisfaction. The robustified controller keeps the entire distribution strictly below the safety threshold (dashed line).}
    \label{fig:boxPlot}
\end{figure}
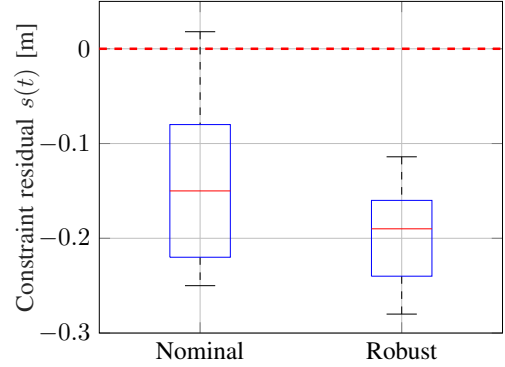

The uncertainty realization includes structured mismatch in mass, inertia, thrust effectiveness, drag, and persistent wind bias, together with a bounded gust component. To keep the numerical assessment compact, two types of evidence are reported: i) a representative closed-loop trajectory comparison between nominal and robustified \ac{NMPC}, and ii) a Monte-Carlo summary over 500 uncertainty realizations.

Figure~\ref{fig:traj_compare} shows a representative inspection run. In the nominal case, the controller tracks the helical reference more aggressively, but under uncertainty the realized trajectory moves closer to the tower. By contrast, the proposed robustified formulation preserves a larger distance from the tower by anticipating the effect of model mismatch and gust disturbances through the tightening margins.

Table~\ref{tab:results_summary} reports the Monte-Carlo results over $500$ realizations with parameters sampled uniformly within the prescribed bounds; ``trials with violation'' counts realizations in which at least one closed-loop instant violates $d_T(\mathbf{p}(t)) \geq d_\mathrm{min}$. Figure~\ref{fig:boxPlot} shows the signed clearance residual $s(t)=d_{\min}-d_T(\mathbf{p}(t))$. The nominal \ac{NMPC} distribution is centered near $-0.15$~m, reflecting the offset between $d_{\mathrm{ref}}=0.35$~m and $d_{\min}=0.20$~m, but the combined effect of wind gusts and structured mismatch yields violations in $230$ of $500$ trials. The robustified \ac{NMPC} shifts the distribution further into the safe region by allocating additional clearance in high-sensitivity portions of the orbit, with median clearance ($\approx0.40$~m) exceeding $d_{\mathrm{ref}}$: the controller prioritizes the tightened safety constraint over reference tracking under uncertainty, as intended. This yields a $100\%$ success rate at the cost of an approximately twofold solve-time increase. The reported MATLAB/MATMPC timings characterize relative overhead rather than embedded real-time feasibility; deployment will rely on a C-code-generated \ac{NMPC} (e.g., via acados~\cite{Verschueren2021}) and is left to future work.

\section{Conclusions}
\label{sec:conclusions}

This paper presented a preliminary sensitivity-based robustification of \ac{NMPC} for close-proximity tower inspection with a tilted multirotor. The method augments the nominal formulation with a parametric tightening term and a stage-dependent gust margin, yielding a conservative clearance constraint under bounded uncertainty. Initial numerical results indicate improved clearance preservation relative to nominal \ac{NMPC}. Future work will focus on broader validation, higher-order sensitivity terms to handle large parametric deviations, and field experiments in a mock-up inspection scenario.



\bibliographystyle{IEEEtran}
\bibliography{references}

\end{document}